\documentclass[10pt,twocolumn,letterpaper]{article}

\usepackage{iccv}
\usepackage{times}
\usepackage{epsfig}
\usepackage{graphicx}
\usepackage{amsmath}
\usepackage{amssymb}

\usepackage{color}
\usepackage{xcolor}
\usepackage{bm}
\usepackage{amsmath}
\usepackage{amssymb}
\usepackage{mathtools}
\usepackage{amsthm}
\usepackage{multirow}
\usepackage{csvsimple}
\usepackage{array}
\usepackage{dblfloatfix}
\usepackage [english]{babel}
\usepackage [autostyle, english = american]{csquotes}
\MakeOuterQuote{"}
\usepackage{subcaption}

\newcommand{\Prob}{\mathbb{P}}
\newcommand{\norm}[1]{\vert \vert#1\vert \vert}
\newcommand{\E}{\mathbb{E}}
\newcommand{\R}{\mathbb{R}}
\newcommand{\D}{\mathcal{D}}
\newcommand{\C}{\mathcal{C}}
\newcommand{\class}[1]{\mathrm{class}(#1)}
\newcommand{\Lwalker}{\mathcal{L_{\mathrm{walker}}}}
\newcommand{\Lvisit}{\mathcal{L_{\mathrm{visit}}}}
\newcommand{\Lclass}{\mathcal{L_{\mathrm{classification}}}}

\newcommand{\Lassoc}{\mathcal{L_{\mathrm{assoc}}}}
\newcommand{\Lsim}{\mathcal{L_{\mathrm{sim}}}}
\newcommand{\DAassoc}{\mathrm{DA}_{\mathrm{assoc}}}
\newcommand{\DAmmd}{\mathrm{DA}_{\mathrm{MMD}}}
\renewcommand{\vec}[1]{\bm{#1}}

\newcommand{\sprod}[2]{\left\langle #1,#2 \right\rangle}
\newcommand{\abs}[1]{\vert #1 \vert}
\newcommand*{\defeq}{\mathrel{\vcenter{\baselineskip0.5ex \lineskiplimit0pt
                     \hbox{\scriptsize.}\hbox{\scriptsize.}}}%
                     =}
                     \definecolor{dgreen}{rgb}{0.0,0.6,0.0}
\definecolor{dred}{rgb}{0.6,0.0,0.0}
\definecolor{darkgreen}{cmyk}{0.90,0.30,0.95,0.30}

\graphicspath{{./figures/}}

\usepackage[pagebackref=true,breaklinks=true,letterpaper=true,colorlinks,bookmarks=false]{hyperref}
\addto\extrasenglish{%
}
\hypersetup{citecolor=red} 
\hypersetup{urlcolor=black}
\hypersetup{linkcolor=dgreen} 
\hypersetup{pdftitle={Associative Domain Adaptation}, %
           pdfauthor={Philip Haeusser, Thomas Frerix, Daniel Cremers}, 
            pdfsubject={ICCV 2017 Submission (arXiv.org version)}, 
            pdfkeywords={deep learning, semi-supervised, training, classification, domain adaptation}} 

\iccvfinalcopy 

\def\iccvPaperID{1052} 

\begin{document}
\title{Associative Domain Adaptation}

\author{Philip Haeusser$^{1,2}$\\
{\tt\small haeusser@in.tum.de}
\and
Thomas Frerix$^1$\\
{\tt\small thomas.frerix@tum.de}
\and
Alexander Mordvintsev$^2$\\
{\tt\small moralex@google.com}
\and
Daniel Cremers$^1$\\
{\tt\small cremers@tum.de}
\and
$^1$Dept. of Informatics, TU Munich\\
\and
$^2$Google, Inc.
}

\maketitle

\begin{abstract}
We propose \emph{associative domain adaptation}, a novel technique for end-to-end domain adaptation with neural networks, the task of inferring class labels for an unlabeled target domain based on the statistical properties of a labeled source domain.
Our training scheme follows the paradigm that in order to effectively derive class labels for the target domain, a network should produce statistically domain invariant embeddings, while minimizing the classification error on the labeled source domain. 
We accomplish this by reinforcing \textnormal{associations} between source and target data directly in embedding space. 
Our method can easily be added to any existing classification network with no structural and almost no computational overhead. 
We demonstrate the effectiveness of our approach on various benchmarks and achieve state-of-the-art results across the board with a generic convolutional neural network architecture not specifically tuned to the respective tasks. Finally, we show that the proposed association loss produces embeddings that are more effective for domain adaptation compared to methods employing maximum mean discrepancy as a similarity measure in embedding space.  
\end{abstract}


\vspace{-\baselineskip}  
\section{Introduction} \label{sec:introduction}
Since the publication of LeNet \cite{LeCun1998} and AlexNet \cite{Krizhevsky2012}, a methodological shift has been observable in the field of computer vision. Deep convolutional neural networks have proved to solve a growing number of problems \cite{szegedy2013deep, Eigen-et-al-14, toshev2014deeppose, szegedy2015going, dosovitskiy2015flownet, mayer2016large}. On the downside, due to a large amount of model parameters, an equally rapidly growing amount of labeled data is needed for training, such as ImageNet \cite{Russakovsky2015}, comprising millions of labeled training examples. This data may be costly to obtain or even nonexistent.

In this paper, we focus on an approach to train neural networks with a minimum of labeled data: domain adaptation.
We refer to domain adaptation as the task to train a model on labeled data from a source domain while minimizing test error on a target domain, for which no labels are available at training time.

\begin{figure}[t!]
	\centering
    \includegraphics[width=1.\linewidth]{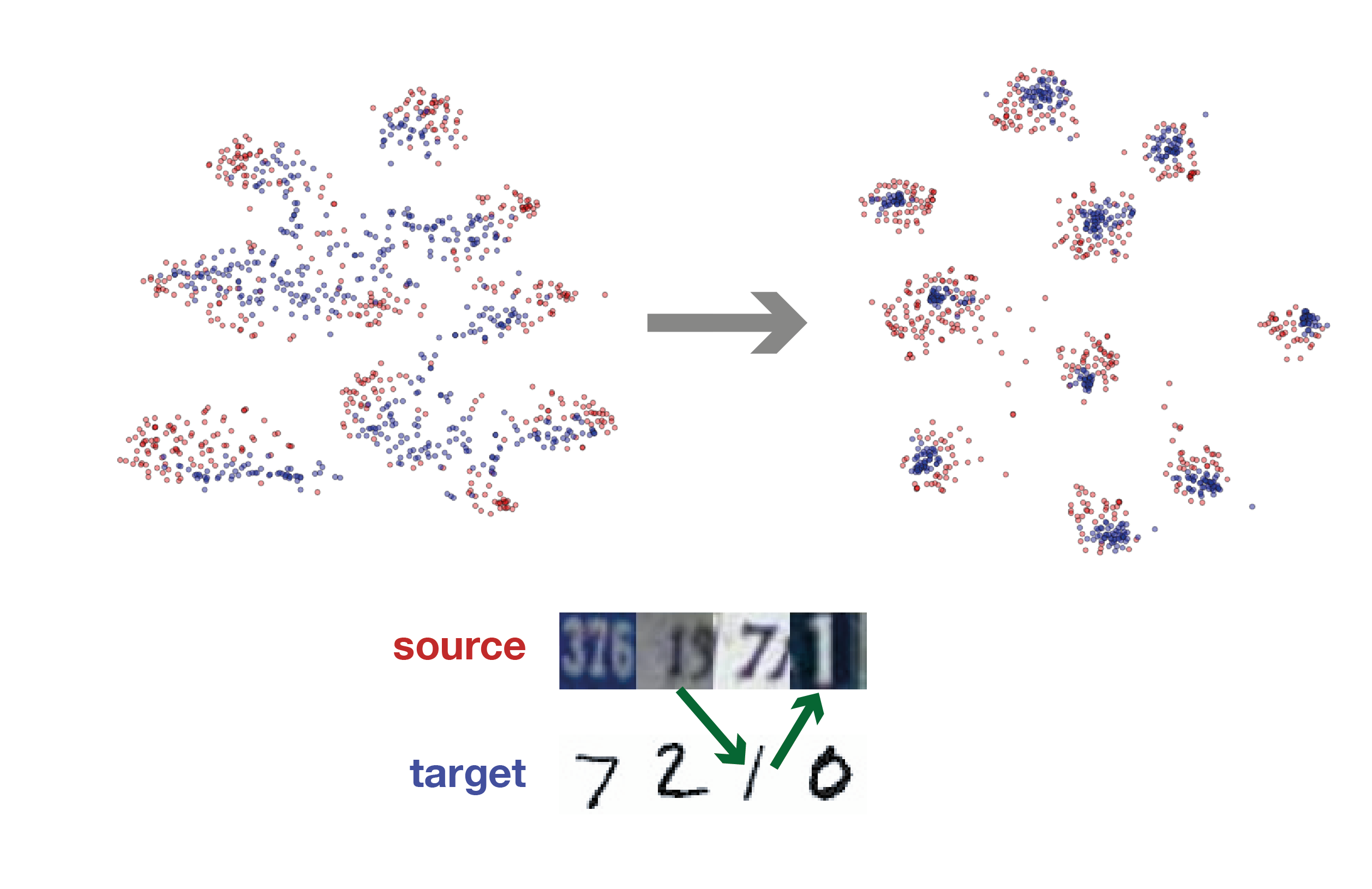}
    \caption{Associative domain adaptation. In order to maximize classification accuracy on an unlabeled target domain, the discrepancy between neural network embeddings of source and target samples (red and blue, respectively) is reduced by an associative loss ($ \color{darkgreen} \rightarrow $), while minimizing a classification error on the labeled source domain.}
	\label{fig:teaser}
\end{figure}

\subsection{Domain adaptation}
\label{sec:domain_adaptation}
In more formal terms, we consider a source domain $\D_s = \{\vec{x}_i^s, y_i^s\}_{i=1,\dots, n_s}$ and a target domain $\D_t~=~\{\vec{x}_i^t, y_i^t\}_{i=1,\dots, n_t}$.
Here, $\vec{x}_i^s \in \R^{N_s}, \vec{x}_i^t \in \R^{N_t}$ are the data vectors and $y_i^s \in \C, y_i^t \in \C$ the respective labels, where the target labels $\{y_i^t\}_{i=1,\dots, n_t}$ are \textit{not} available for training.
Note that for domain adaption it is assumed that source and target domains are associated with the same label space, while $\D_s$ and $\D_t$ are drawn from distributions $\Prob_s$ and $\Prob_t$, which are assumed to be \textit{different}, i.e. the source and target distribution have different joint distributions of data $\vec{X}$ and labels $\vec{Y}$, $\Prob_s(\vec{X}, \vec{Y}) \neq\Prob_t(\vec{X}, \vec{Y})$.

The value of domain adaptation has even more increased with generative tools producing synthetic datasets.
The idea is compelling: rather than labeling vast amounts of real-world data, one renders a similar but synthetic dataset that is automatically labeled. With an effective method for domain adaptation it becomes possible to train models without the need for one single labeled target  example at training time. 

In order to combine labeled and unlabeled data for a predictive task, a variety of notions has emerged.  
To be clear, we explicitly distinguish \emph{domain adaptation} from related approaches.
For semi-supervised learning, labeled source data is leveraged by unlabeled target data drawn from the \textit{same} distribution, i.e. $\Prob_s = \Prob_t$.
In transfer learning, not only source and target domain are drawn from different distributions, also their label spaces are generally different.
An example of supervised transfer learning is training a neural network on a source domain and subsequently fine-tuning the model on a labeled target domain for a different task \cite{Yosinski2014, Donahue2014}.

The problem of domain adaptation was theoretically studied in \cite{Ben-David2010}, relating source and target error with a statistical similarity measure of the respective domains.
Their results suggest that a good domain adaptation method should be based on features that are as similar as possible for source and target domain (\emph{assimilation}), while reducing the prediction error in the source domain as much as possible (\emph{discrimination}).
These effects are opposing each other since source and target domains are drawn from different distributions. This can be formulated as a cost function that consists of two terms:

\begin{equation}
    \mathcal{L} = \Lclass + \Lsim \; ,
    \label{eq:domain_adaptation_formalism}
\end{equation}

Here, the classification loss, $\Lclass$ encourages discrimination between different classes, maximizing the margin between clusters of embeddings that belong to the same class.
We define the second term as a generic similarity loss $\Lsim$, which enforces statistically similar latent representations.

Intuitively, for similar latent representations of the source and target domain, the target class labels can be more accurately inferred from the labeled source samples.

In the following, we show how previous methods approached this optimization and then propose a new loss for $\Lsim$.

\subsection{Related work} 
\label{sec:related_work}
Several works have approached the problem of domain adaptation.
Here, we mainly focus on methods that are based on deep learning, as these have proved to be powerful learning systems and are closest to our scheme.

The CORAL method \cite{Sun2016b} explicitly forces the covariance of the target data onto the source data (\emph{assimilation}).
The authors then apply supervised training to this transformed source domain with original labels (\emph{discrimination}).
This idea is extended to second order statistics of features in deep neural networks in \cite{Sun2016}.

Building on the idea of adversarial training \cite{Goodfellow2014a}, the authors of \cite{Ganin2015a} propose an architecture in which a class label and a domain label predictor are built on top of a general feature extractor. 
While the class label predictor is supposed to correctly classify the labeled training examples (\emph{discrimination}), the domain label predictor for all training samples is used in a way to make the feature distributions similar (\emph{assimilation}).
The authors of \cite{Bousmalis2016b} use an adversarial approach to train for similarity in data space instead of feature space.
Their training scheme is closer to standard generative adversarial networks \cite{Goodfellow2014a}, however, it does not only condition on noise, but also on an image from the source domain.
 
Within the paradigm of training for domain invariant features, one popular metric is the maximum mean discrepancy (MMD) \cite{Gretton2012}.
This measure is the distance between the mean embeddings of two probability distributions in a reproducing kernel Hilbert space $\mathcal{H}_k$ with a characteristic kernel $k$.
More precisely, the mean embedding of a distribution $\Prob$ in $\mathcal{H}_k$ is the unique element $\mu_k(\Prob) \in \mathcal{H}_k$ such that $\E_{x \sim \Prob}[f(x)] =  \sprod{f(x)}{\mu_k(\Prob)}_{\mathcal{H}_k}, \forall f \in \mathcal{H}_k$.
The MMD distance between source and target domain then reads $d_{\mathrm{MMD}}(\Prob_s, \Prob_t) = \norm{\mu_k(\Prob_s) - \mu_k(\Prob_t)}_{\mathcal{H}_k}$.
In practice, this distance is computed via the kernel trick \cite{Vapnik1995}, which leads to an algorithm with quadratic runtime in the number of samples.
Linear time estimators have previously been proposed \cite{Long2015a}.

Most works, which explicitly minimize latent feature discrepancy, use MMD in some variant. That is, they use MMD as $\Lsim$ in order to achieve \emph{assimilation} as defined above.
The authors of \cite{Long2015a} propose the Deep Adaptation Network architecture.
Exploiting that learned features transition from general to specific within the network, they train the first layers of a CNN commonly for source and target domain, then train individual task-specific layers while minimizing the multiple kernel maximum mean discrepancies between these layers. 

The technique of task-specific but coupled layers is further explored in \cite{Rozantsev2016} and \cite{Bousmalis2016}.
The authors of \cite{Rozantsev2016} propose to individually train source and target domains while the network parameters of each layer are regularized to be linear transformations of each other.
In order to train for domain invariant features, they minimize the MMD of the embedding layer.
On the other hand, the authors of \cite{Bousmalis2016} maintain a shared representation of both domains and private representations of each individual domain in their Domain Separation architecture.

As becomes evident in these works, the MMD minimizes domain discrepancy in some abstract space and requires a choice of kernels with appropriate hyperparameters, such as the standard deviation of the Gaussian kernel. In this work, we propose a different loss for $\Lsim$ which is more intuitive in embedding space, less computationally complex and better suitable to obtain effective embeddings.

\subsection{Contribution}
We propose the association loss $\Lassoc$ as an alternative discrepancy measure ($\Lsim$) within the domain adaptation paradigm described in \autoref{sec:domain_adaptation}.
The reasoning behind our approach is the following: Ultimately, we want to minimize the classification error on the target domain $\D_t$. This is not directly possible since no labels are available at training time. Therefore, we minimize the classification error on the source domain $\D_s$ as a proxy while enforcing representations of $\D_t$ to have similar statistics to those of $\D_s$. This is accomplished by enforcing \emph{associations} \cite{Haeusser2017} between feature representations of $\D_t$ with those of $\D_s$ that are in the same class.
Therefore, in contrast to MMD as $\Lsim$, this approach also leverages knowledge about labels of the source domain and hence avoids unwanted \emph{assimilation} across class clusters.
The implementation is simple yet powerful as we show in \autoref{sec:domain_adaptation_by_associative_learning}. It works with any existing architecture and, unlike most deep learning approaches for domain adaptation, does not introduce a structural and almost no computational overhead. 
In fact, we used the same generic and simple architecture for \emph{all} our experiments, each of which achieved state-of-the-art results.

In summary, our contributions are:
\begin{itemize}
    \item A straightforward training schedule for domain adaptation with neural networks. 
    \item An integration of our approach into the prevailing domain adaptation formalism and a detailed comparison with the most commonly used explicit $\Lsim$: the maximum mean discrepancy (MMD).    
    \item A simple implementation that works with arbitrary architectures\footnote{\url{https://git.io/vyzrl}}.
    \item Extensive experiments on various benchmarks for domain adaptation that outperform related deep learning methods.
    \item A detailed analysis demonstrating that associative domain adaptation results in effective embeddings in terms of classifying target domain samples.
\end{itemize}

\section{Associative domain adaptation}\label{sec:domain_adaptation_by_associative_learning}
We start from the approach of learning by association \cite{Haeusser2017} which is geared towards semi-supervised training. 
Labeled and unlabeled data are related by associating their embeddings, i.e. features of a neural network's last layer before the softmax layer.
Our work generalizes this approach for domain adaptation.
For the new task, we identify labeled data with the source domain and unlabeled data with the target domain.
Specifically, for $\vec{x}_i^s \in \D_s, \vec{x}_i^t \in \D_t$ and the embedding map $\phi : \mathbb{R}^{N_0} \rightarrow \mathbb{R} ^{N_{L-1}}$ of an $L$-layer neural network, denote by $A_i \defeq \phi(\vec{x}_i^s), B_j \defeq \phi(\vec{x}_j^t)$ the respective embeddings of source and target domain.
Then, similarity is measured by the embedding vectors' dot product as $M_{ij} = \sprod{A_i}{B_j}$.

If one considers transitions between the parts $(\{A_i\}, \{B_j\})$ of a bipartite graph, the intuition is that transitions are more probable if embeddings are more similar. This is formalized by the transition probability from embedding $A_i$ to embedding $B_j$:
\begin{equation}
	P_{ij}^{ab} = \Prob(B_j \vert A_i) \defeq \frac{\exp(M_{ij})}{\sum_{j'}\exp(M_{ij'})} \; .
	\label{eq:softmax_transitions}
\end{equation}

The basis of associative similarity is the two-step round-trip probability of an imaginary random walker starting from an embedding $A_i$ of the labeled source domain and returning to another embedding $A_j$ via the (unlabeled) target domain embeddings $B$,

\begin{equation}
	P_{ij}^{aba} \defeq \left(P^{ab} P^{ba}\right)_{ij} \;.
	\label{eq:def_roundtrip}
\end{equation}

The authors of \cite{Haeusser2017} observed that higher order round trips do not improve performance.
The two-step probabilities are forced to be similar to the uniform distribution over the class labels via a cross-entropy loss term called the \textit{walker loss},
\begin{equation}
	\Lwalker \defeq H\left(T, P^{aba}\right) \; ,
	\label{eq:def_walker_loo}
\end{equation}
where
\begin{align}
	T_{ij} \defeq 
	\begin{cases}
		1 / \abs{A_i}  &\class{A_i} = \class{A_j} \\
		0 &\mathrm{else}
	\end{cases}
	\label{eq:def_uniform_class}
\end{align}

This means that all association cycles within the same class are forced to have equal probability. The walker loss by itself could be minimized by only visiting target samples that are easily associated, skipping difficult examples. This would lead to poor generalization to the target domain.
Therefore, a regularizer is necessary such that each target sample is visited with equal probability.
This is the function of the \textit{visit loss}. It is defined by the cross entropy between the uniform distribution over target samples and the probability of visiting some target sample starting in any source sample,
\begin{equation}
	\Lvisit \defeq H(V, P^{\mathrm{visit}})	\; ,
	\label{eq:def_visit_loss}
\end{equation}
where 
\begin{equation}
	P_j^{\mathrm{visit}} \defeq \sum_{\vec{x}_i \in \D_s} P_{ij}^{ab} ,\quad V_j \defeq \frac{1}{\abs{B}} \, .
	\label{eq:def_visit_prob}
\end{equation}

Note that this formulation assumes that the class distribution is the same for source and target domain. If this is not the case, using a low weight for $\Lvisit$ may yield better results.

Together, these terms form a loss that enforces associations between similar embeddings of both domains,  
\begin{equation}
    \Lassoc = \beta_1 \Lwalker + \beta_2 \Lvisit \; ,
    \label{eq:association_loss}
\end{equation}
where $\beta_i$ is a weight factor.
At the same time, the network is trained to minimize the prediction error on the labeled source data via a softmax cross-entropy loss term, $\Lclass$.

The overall neural network loss for our training scheme is given by
\begin{equation}
	\mathcal{L} = \Lclass + \alpha \Lassoc \; .
	\label{eq:total_loss}
\end{equation}

We want to emphasize once more the essential motivation for our approach: The association loss enforces similar embeddings (\emph{assimilation}) for the source and target samples, while the classification loss minimizes the prediction error of the source data (\emph{discrimination}).
Without $\Lassoc$, we have the case of a neural network that is trained conventionally \cite{Krizhevsky2012} on the source domain only. 
As we show in this work, the (scheduled) addition of $\Lassoc$ during training allows to incorporate unlabeled data from a different domain improving the effectiveness of embeddings for classification.
Adding $\Lassoc$ enables an arbitrary neural network to be trained for domain adaptation.
The neural network learning algorithm is then able to model the shift in distribution between source and target domain.
More formally, if $\Lassoc$ is minimized, \emph{associated} embeddings from both source and target domain become more similar in terms of their dot product. 

In contrast to MMD, $\Lassoc$ incorporates knowledge about source domain classes and hence prevents the case that source and target domain embeddings are statistically similar, but not class discriminative. We demonstrate this experimentally in \autoref{sec:analysis}.

We emphasize that not every semi-supervised training method can be adapted for domain adaptation in this manner. It is necessary that the method explicitly models the shift between the source and target distributions, in order to reduce the discrepancy between both domains, which is accomplished by $\Lassoc$.

In this respect, associative domain adaptation parallels the approaches mentioned in \autoref{sec:related_work}. As we demonstrate experimentally in the next section, $\Lassoc$ is employed as a compact, intuitive and effective training signal for \emph{assimilation} yielding superior performance on all tested benchmarks.

\section{Experiments} 

\begin{table}[t!]
    \centering
    \begin{tabular}{>{\centering\arraybackslash}m{2cm}  >{\centering\arraybackslash}m{4cm}}
        \shortstack{\textsc{MNIST} \\$\big\Downarrow$\\ \textsc{MNIST-M}\\\small(10 classes)}  & \includegraphics[width=3cm]{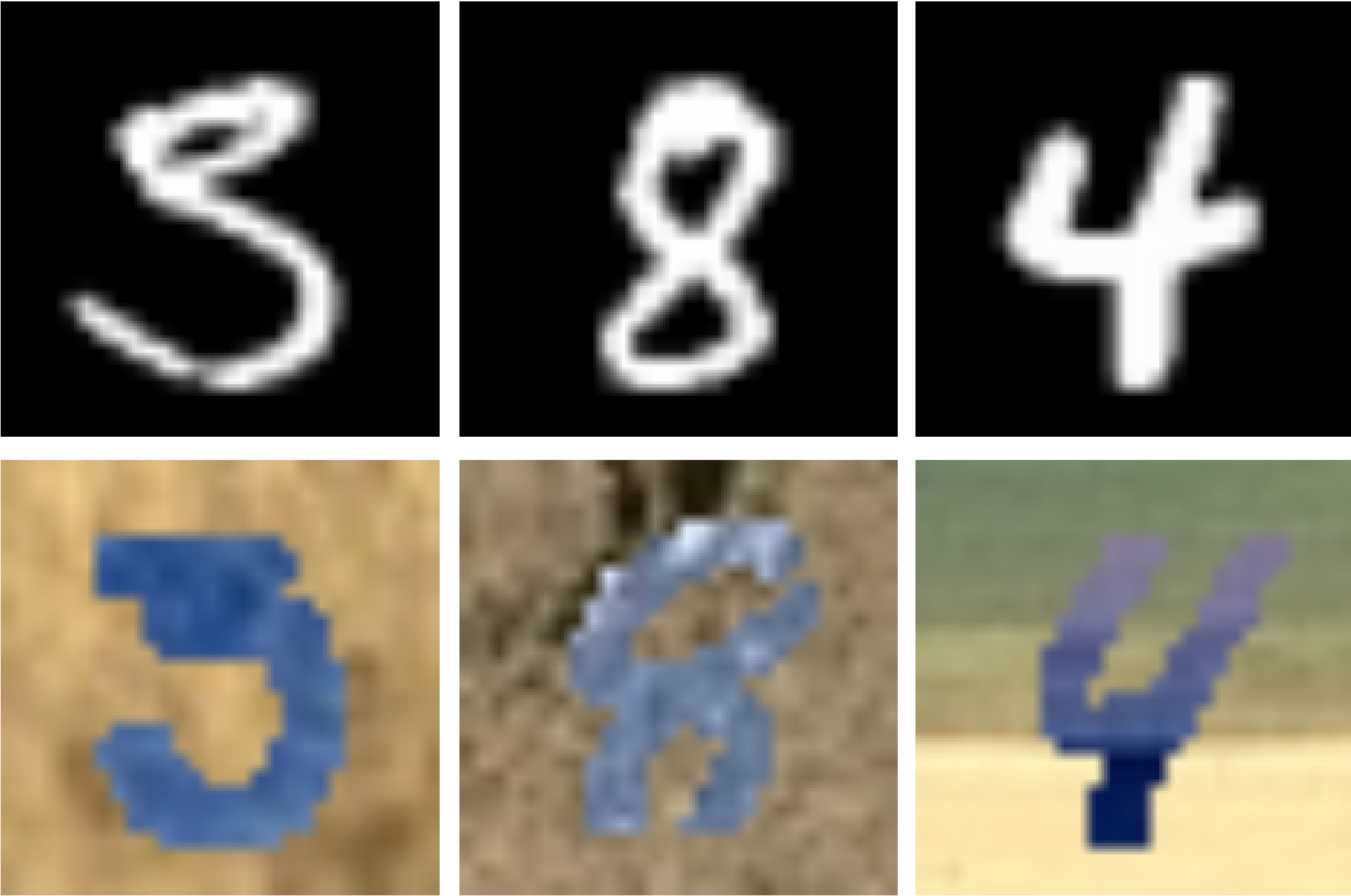} \\[0.3cm] 
        \shortstack{\textsc{Synth} \\$\big\Downarrow$\\ \textsc{SVHN}\\\small(10 classes)}  & \includegraphics[width=3cm]{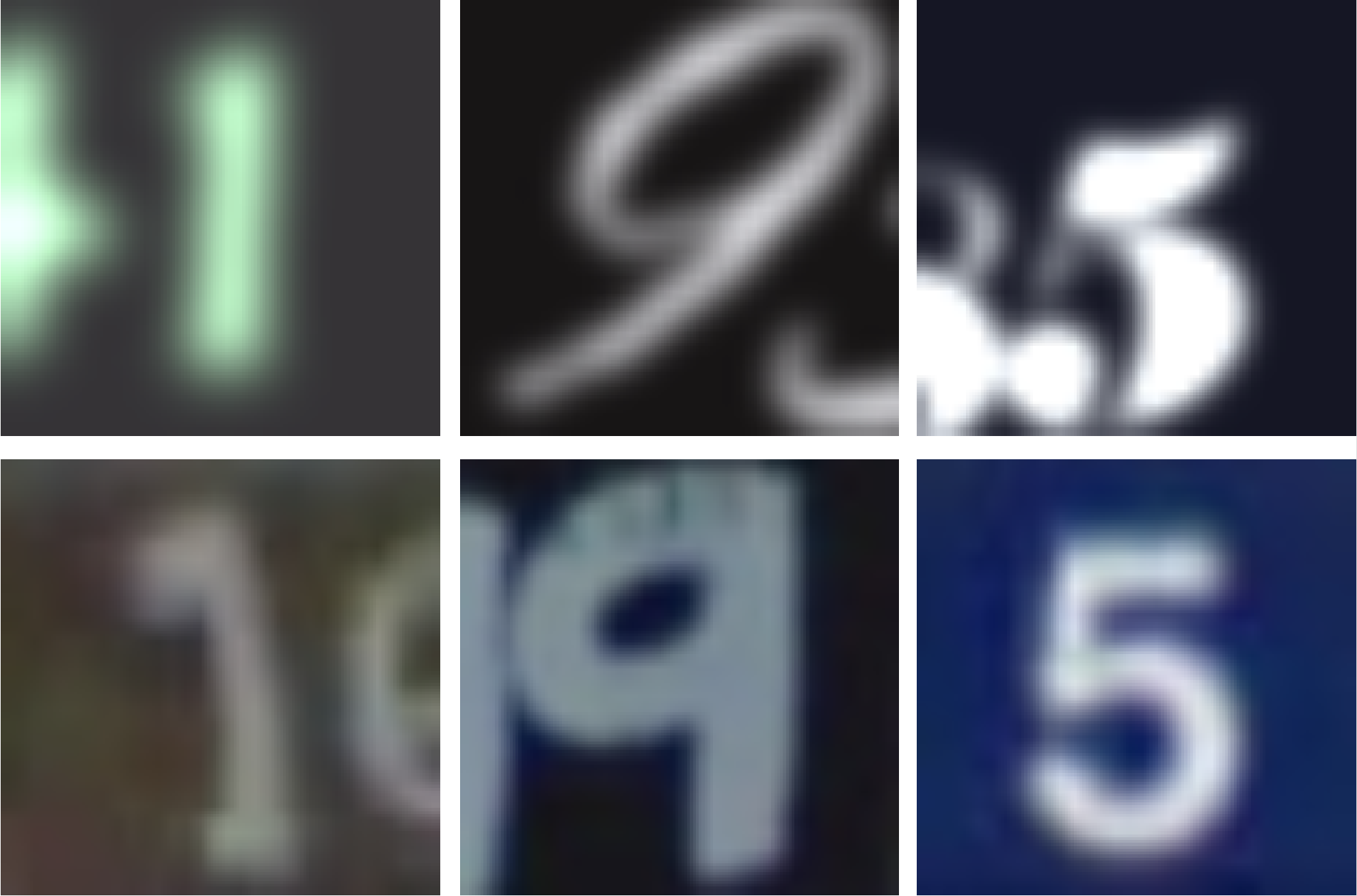} \\ [0.3cm] 
        \shortstack{\textsc{SVHN} \\$\big\Downarrow$\\ \textsc{MNIST}\\\small(10 classes)}  & \includegraphics[width=3cm]{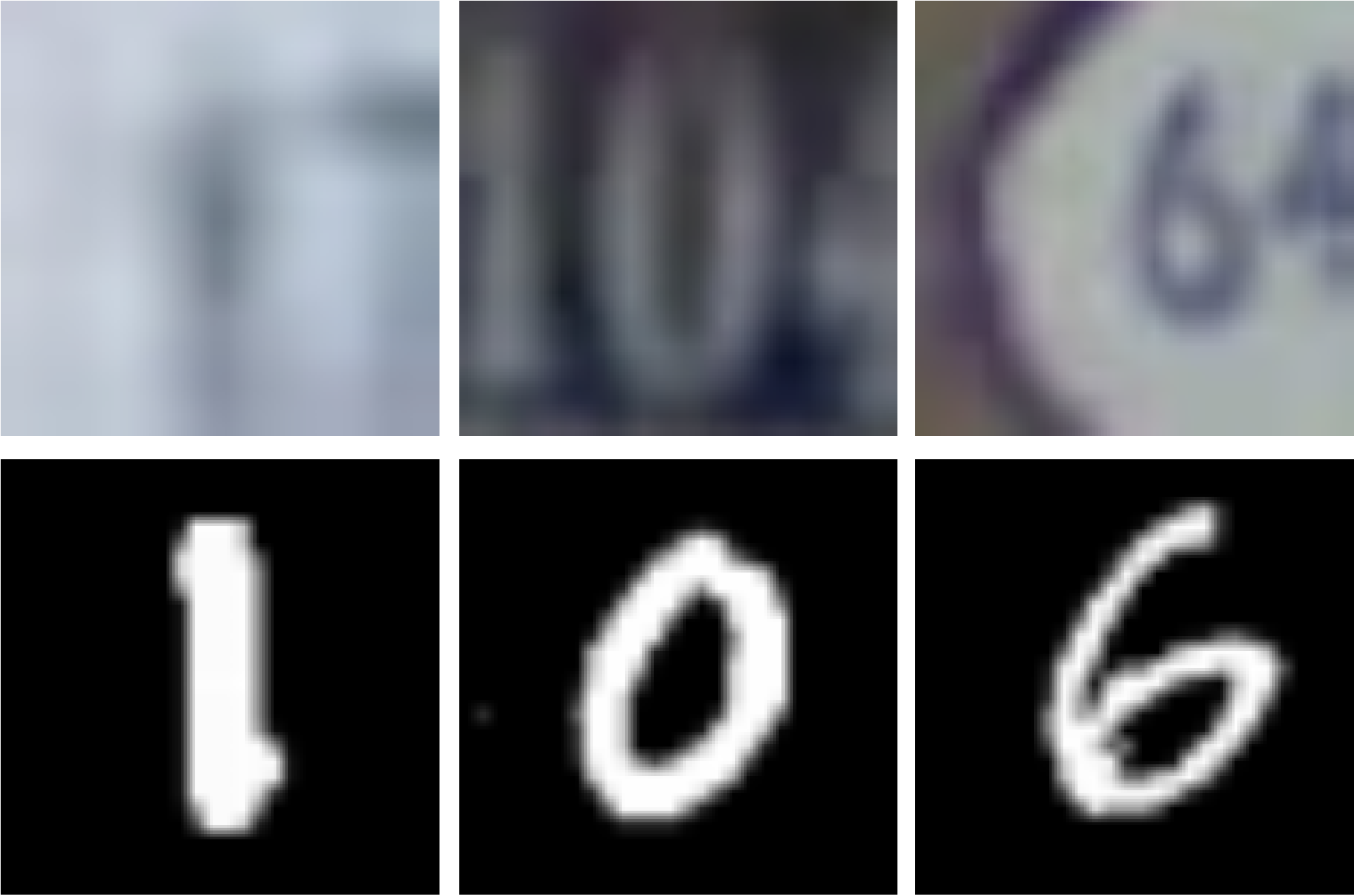} \\ [0.3cm] 
        \shortstack{\textsc{Synth Signs} \\$\big\Downarrow$\\ \textsc{GTSRB}\\\small(43 classes)}  & \includegraphics[width=3cm]{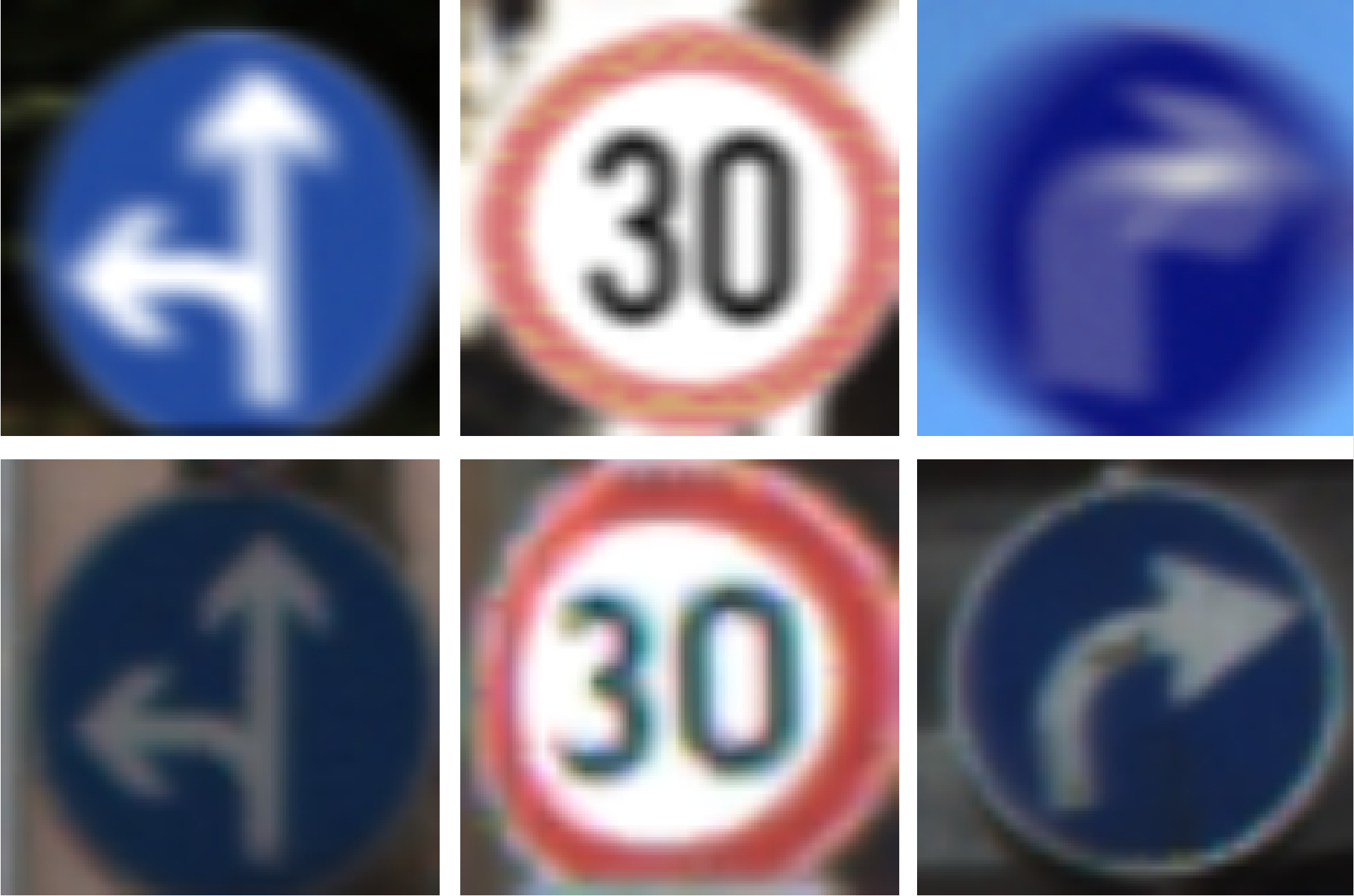} 
    \end{tabular}
    \caption{Dataset samples for our domain adaptation tasks. For three randomly chosen classes, the first row depicts a source sample, the second row a target sample. The datasets vary in difficulty due to differences in color space, variance of transformation or number of classes. }
    \label{tab:dataset_juxtaposition}
\end{table}

\begin{table*}[!b]
	\begin{center}
		\begin{tabular}{|c||c|c|c|c|}
			\hline
			\multirow{ 2}{*}{Method} & \multicolumn{4}{c|}{Domains (source $\rightarrow$ target) } \\
			& MNIST $\rightarrow$ MNIST-M
			& Syn. Digits $\rightarrow$ SVHN 
			& SVHN $\rightarrow$ MNIST
			& Syn. Signs $\rightarrow$ GTSRB\\
			\hline
			\hline
            Transf. Repr. \cite{Sener2016} & 13.30 & - & 21.20 & -\\
            SA \cite{Fernando2013} & 43.10 & 13.56 & 40.68 & 18.35\\
            CORAL \cite{Sun2016b} & 42.30 & 14.80 & 36.90 & 13.10\\
            ADDA \cite{Tzeng2016} & - & - & 24.00 & -\\
            DANN \cite{Ganin2015a} & 23.33 (55.87 \%) & 8.91 (79.67 \%) & 26.15 (42.57 \%) & 11.35 (46.39 \%)\\
            DSN w/ DANN \cite{Bousmalis2016b} & 16.80 (63.18 \%) & 8.80 (78.95 \%) & 17.30 (58.31 \%) & 6.90 (54.42 \%)\\
            DSN w/ MMD \cite{Bousmalis2016b} & 19.50 (56.77 \%) & 11.50 (31.58 \%) & 27.80 (32.26 \%) & 7.40 (51.02 \%)\\
            MMD \cite{Long2015a} & 23.10 & 12.00 & 28.90 & 8.90\\
            \hline
            $\DAmmd$ & 22.90 & 19.14 & 28.48 & 10.69\\
            Ours ($\DAassoc$ fixed params$^\dagger$) & \bf 10.47 $\pm$ 0.28 & 8.70 $\pm$ 0.2 & 4.32 $\pm$ 1.54 & 17.20 $\pm$ 1.32\\
            \bf Ours ($\DAassoc$) & 10.53 (85.94 \%) & \bf 8.14 (87.78 \%) & \bf 2.40 (93.71 \%) & \bf 2.34 (81.23)\\
            \hline
            Source only & 35.96 & 15.68 & 30.71 & 4.59\\
            Target only & 6.37 & 7.09 & 0.50 & 1.82\\
			\hline
		\end{tabular}
	\end{center}
\caption{
Domain adaptation. Errors (\%) on the target test sets (lower is better).
\textit{Source only} and \textit{target only} refer to training only on the respective dataset (supervisedly \cite{Haeusser2017}, without domain adaptation) and evaluating on the target dataset.
In the $\DAmmd$ setting, we replaced $\Lassoc$ with MMD.
The metric \emph{coverage} is reported in parentheses, where available (cf. \autoref{eqn:coverage}). We used the same network architecture for all our experiments and achieve state of the art results on all benchmarks. The row "$\DAassoc$ fixed params$^\dagger$" reports results from 10 runs ($\pm$ standard deviation) with an arbitrary choice of fixed hyper parameters ($\beta_2$ = 0.5, delay = 500 steps and batch size = 100) for all four domain pairs. The row below shows our results after individual hyper parameter optimization.
No labels of the target domain were used at training time.}
\label{tbl:da}
\end{table*}

\subsection{Domain adaptation benchmarks}
In order to evaluate and compare our method, we chose common domain adaptation tasks, for which previous results are reported. Examples for the respective datasets are shown in \autoref{tab:dataset_juxtaposition}.

\paragraph{MNIST $\rightarrow$ MNIST-M}
We used the MNIST \cite{LeCun1998} dataset as labeled source and generated the unlabeled MNIST-M target as described in \cite{Ganin2015a}.
Background patches from the color photo BSDS500 dataset \cite{Arbelaez2011} were randomly extracted. Then the absolute value of the difference of each color channel with the MNIST image was taken.
This yields a color image, which can be easily identified by a human, but is significantly more difficult for a machine compared to MNIST due to two additional color channels and more nuanced noise.
The single channel of the MNIST images was replicated three times to match those of the MNIST-M images (RGB). The image size is $28 \times 28$ pixels. This is the only setting where we used data augmentation: We randomly inverted MNIST images since they are always white on black, unlike MNIST-M.

\paragraph{Synth $\rightarrow$ SVHN}
The Street View House Numbers (SVHN) dataset \cite{Netzer2011} contains house number signs extracted from Google Street View. We used the variant \emph{Format 2} where images ($32 \times 32$ pixels) are already cropped. Still, multiple digits can appear in one image. As a labeled source domain we use the Synthetic Digits dataset provided by the authors of \cite{Ganin2015a}, which expresses a varying number of fonts and properties (background, orientation, position, stroke color, blur) that aim to mimic the distribution in SVHN.

\paragraph{SVHN $\rightarrow$ MNIST}
MNIST images were resized with bilinear interpolation to $32 \times 32$ pixels and extended to three channels in order to match the shape of SVHN.

\paragraph{Synthetic Signs $\rightarrow$ GTSRB}
The Synthetic Signs dataset was provided by the authors of \cite{Moiseev2013} and consists of 100,000 images that were generated by taking common street signs from Wikipedia and applying various artificial transformations.
The German Traffic Signs Recognition Benchmark (GTSRB) \cite{Stallkamp2011} provides 39,209 (training set) and 12,630 (test set) cropped images of German traffic signs.
The images vary in size and were resized with bilinear interpolation to match the Synthetic Signs images' size of $40 \times 40$ pixels. Both datasets contain images from 43 different classes.

\subsection{Training setup}\label{sec:setup}
\subsubsection{Associative domain adaptation}
Our formulation of \emph{associative domain adaptation} is implemented\footnote{\url{https://git.io/vyzrl}} as a custom loss function that can be added to any existing neural network architecture.
Results obtained by neural network learning algorithms often highly depend on the complexity of a specifically tuned architecture.
Since we wanted to make the effect of our approach as transparent as possible, we chose the following generic convolutional neural network architecture for \emph{all} our experiments:
\begin{align*}
&C(32,3)\rightarrow C(32, 3)\rightarrow P(2)\\
\rightarrow \; &C(64, 3)\rightarrow C(64, 3)\rightarrow P(2)\\
\rightarrow\;  &C(128, 3)\rightarrow C(128, 3)\rightarrow P(2)\rightarrow FC(128)
\end{align*}

Here, $C(n, k)$ stands for a convolutional layer with n kernels of size $k \times k$ and stride 1.
$P(k)$ denotes a max-pooling layer with window size $k \times k$ and stride 1.
$FC(n)$ is a fully connected layer with $n$ output units. The size of the embeddings is 128. An additional fully connected layer maps these embeddings to logits, which are the input to a softmax cross-entropy loss for classification, $\Lclass$.

The detailed hyperparameters for each experiment can be found in the supplementary material. The most important hyperparameters are the following:
\paragraph{Learning rate}
We chose the same initial learning rate ($\tau = 1e^{-4}$) for all experiments, which was reduced by a factor of $0.33$ in the last third of the training time. All trainings converged in less than 20k iterations.

\paragraph{Mini-batch sizes}
It is important to ensure that a mini-batch represents all classes sufficiently, in order not to introduce a bias. 
For the labeled mini-batch, we explicitly sample a number of examples per class. 
For the unlabeled mini-batch we chose the same overall size as for the labeled one, usually around 10-100 times the number of classes.

\paragraph{Loss weights}
The only loss weight that we actively chose is the one for $\Lvisit$, $\beta_2$. As was shown in \cite{Haeusser2017}, this loss acts as a regularizer. Since it assumes the same class distribution on both domains, the weight needs to be lowered if the assumption does not hold. We experimentally chose a suitable weight.

\paragraph{Delay of $\Lassoc$}
We observed that convergence is faster if we first train the network only with the classification loss, $\Lclass$, and then add the association loss, $\Lassoc$, after a number of iterations. 
This is implemented by defining $\alpha$ (\autoref{eq:association_loss}) as a step function.
This procedure is intuitive, as the transfer of label information from source to target domain is most effective when the network has already learned some class structure and the embeddings are not random anymore.

\begin{figure*}[htb]
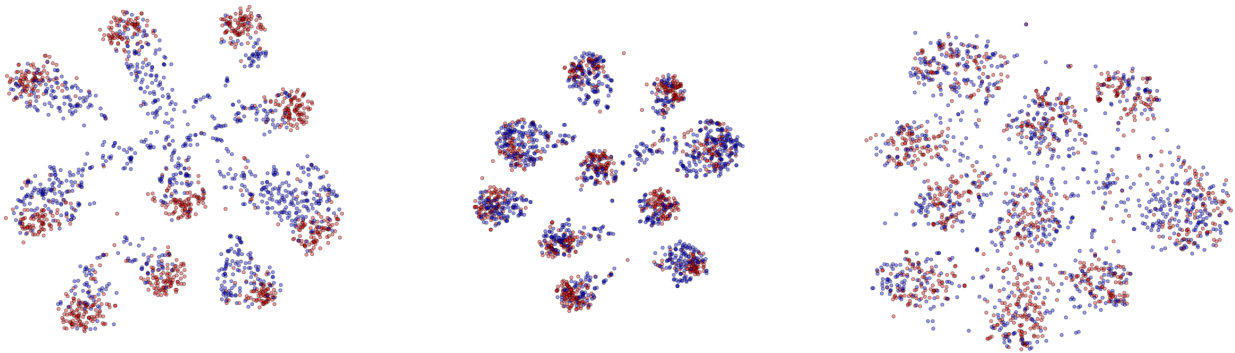

	\centering
	    \begin{subfigure}{0.33\textwidth}
        \includegraphics[width=\textwidth]{tsne_synth2svhn_SO.png}
        \label{fig:tsne_source_only}
    \end{subfigure}%
    \begin{subfigure}{0.33\textwidth}
        \includegraphics[width=\textwidth]{tsne_synth2svhn_DA.png}
        \label{fig:tsne_domain_adaptation}
    \end{subfigure}%
    \begin{subfigure}{0.33\textwidth}
        \includegraphics[width=\textwidth]{tsne_synth2svhn_MMD.png}
        \label{fig:tsne_mmd}
    \end{subfigure}%
	\caption{t-SNE embeddings with perplexity $35$ of 1,000 test samples for Synthetic Digits (source, red) and SVHN (target, blue).
	\textbf{Left}: After training on \emph{source only}. \textbf{Middle}: after training with \emph{associative domain adaptation} ($\DAassoc$). \textbf{Right}: after training with \emph{MMD loss} ($\DAmmd$). While the target samples are diffuse when embedded with the \emph{source only} trained network, the class label information is successfully inferred after \emph{associative domain adaptation}. When the network is trained with an \textit{MMD loss}, the resulting distributions are similar, but less visibly class discriminative.}
	\label{fig:tsne}
\end{figure*}
\begin{table*}[htb]
	\begin{center}
		\begin{tabular}{|c||c|c|c|c|}
			\hline
			\multirow{ 2}{*}{} & \multicolumn{4}{c|}{Domains (source $\rightarrow$ target) } \\
			& MNIST $\rightarrow$ MNIST-M
			& Syn. Digits $\rightarrow$ SVHN 
			& SVHN $\rightarrow$ MNIST
			& Syn. Signs $\rightarrow$ GTSRB\\
			\hline
            Source only             & 0.1234 (35.96) & 0.1010 (15.68) & 0.0739 (30.71) & 0.0466 (4.59)\\
            $\DAassoc$   & 0.0504 (10.53) & 0.0415 (8.14) & 0.2112 (2.40) & 0.0459 (2.34)\\            
            $\DAmmd$         & 0.0233 (22.90) & 0.0166 (19.29) & 0.0404 (34.06) & 0.0145 (12.85) \\
			\hline
		\end{tabular}
	\end{center}
\caption{
Maximum mean discrepancy (MMD) between embeddings of source and target domain, obtained with a network trained supervisedly on source only (SO), for the domain adaptation setting with $\Lassoc$ ($\DAassoc$) and with an MMD loss ($\DAmmd$). Numbers in parentheses are test errors on the target domain from \autoref{tbl:da}. Associative domain adaptation also reduces the MMD in some cases. Lower MMD values do not correlate with lower test errors. In fact, even though the MMD for training with the associative loss is higher compared with training with the MMD loss, our approach achieves lower test errors.}
\label{tbl:mmd}
\end{table*}

\paragraph{Hyper parameter tuning}
We are aware that hyper parameter tuning can sometimes obscure the actual effect of a proposed method. In particular, we want to discuss the effect of small batch sizes on our algorithm. For the association loss to work properly, all classes must be represented in a mini-batch, which places a restriction on small batch sizes, when the number of classes is large. To further investigate this hyperparameter we ran the same architecture with an arbitrary choice of fixed hyper parameters and smaller batch size ($\beta_2$ = 0.5, delay = 500 steps and batch size = 100) for all four domain pairs and report the mean and standard deviation of 10 runs in the row "$\DAassoc$ fixed params$^\dagger$". In all cases except for the traffic signs, these runs outperform previous methods. The traffic sign setup is special because there are 4.3$\times$ more classes and with larger batches more classes are expected to be present in the unlabeled batch. When we removed the batch size constraint, we achieved a test error of 6.55 $\pm$ 0.59, which outperforms state of the art for the traffic signs.

\paragraph{Hardware}
All experiments were carried out on an NVIDIA Titan X (Pascal). Each experiment took less than 120 minutes until convergence.

\subsubsection{Domain adaptation with MMD}\label{sec:mmdsetup}
In order to compare our setup and the proposed $\Lassoc$, we additionally ran all experiments described above with MMD instead of $\Lassoc$. We performed the same hyperparameter search for $\alpha$ and report the respectively best test errors. 
We used the open source implementation
including hyperparameters from \cite{sutherland2016generative}.
This setup is referred to as $\DAmmd$.

\subsection{Evaluation}
All reported test errors are evaluated on the target domain.
To assess the quality of domain adaptation, we provide results trained on source and target only (SO and TO, respectively) as in \cite{Haeusser2017}, for associative domain adaptation ($\DAassoc$) and for the same architecture with MMD instead of $\Lassoc$.
Besides the absolute accuracy, an informative metric is \emph{coverage} of the gap between TO and SO by DA,

\begin{equation*}\label{eqn:coverage}
    \frac{DA - SO}{TO - SO} \; ,
\end{equation*}
as it is a measure of how much label information is successfully transferred from the source to the target domain. In order to assess a method's performance on domain adaptation, one should always consider both coverage and absolute error on the target test set since a high coverage could also stem from poor performance in the SO or TO setting.

Where available, we report the coverage of other methods (with respect to their own performance on SO and TO).

\autoref{tbl:da} shows the results of our experiments. In all four popular domain adaptation settings our method performs best. On average, our approach improves the performance by 87.17~\% compared to training on source only (\emph{coverage}). In order to make our results as comparable as possible, we used a generic architecture that was not handcrafted for the respective tasks (cf. \autoref{sec:setup}).

\subsection{Analysis of the embedding quality}
\label{sec:analysis}
As described in \autoref{sec:introduction}, a good intuition for the formalism of domain adaptation is the following.
On the one hand, the latent features should cluster in embedding space, if they belong to the same class (\emph{assimilation}). 
On the other hand, these clusters should separate well in order to facilitate classification (\emph{discrimination}). 

We claim that our proposed $\Lassoc$ is well suited for this task compared with maximum mean discrepancy. We use four points to support this claim:

\begin{itemize}
    \item t-SNE visualizations show that employing $\Lassoc$ produces embeddings that cluster better compared to MMD.
    \item $\Lassoc$ simultaneously reduces the maximum mean discrepancy (MMD) in most cases.
    \item Lower MMD values do not imply lower target test errors in these settings.
    \item In all cases, the target domain test error of our approach is lower compared to training with an MMD loss.
\end{itemize}

\subsubsection{Qualitative evaluation: t-SNE embeddings}\label{sec:tsne}
A popular method to visualize high-dimensional data in 2D is t-SNE \cite{maaten2008visualizing}. 
We are interested in the distribution of embeddings for source and target domain when we employ our training scheme. 
\autoref{fig:tsne} shows such visualizations. We always plotted embeddings of the target domain test set.
The embeddings are obtained with networks trained semi-supervisedly \cite{Haeusser2017} on the source domain only (SO), with our proposed associative domain adaptation ($\DAassoc$) and with MMD instead of $\Lassoc$ ($\DAmmd$, cf. \autoref{sec:setup}).

In the SO setting, samples from the source domain fall into clusters as expected. Samples from the target domain are more scattered.
For $\DAassoc$, samples from both domains cluster well and become separable.
For $\DAmmd$, the resulting distributions are similar, but not visibly class discriminative. 

For completeness, however, we explicitly mention that t-SNE embeddings are obtained via a non-linear, stochastic optimization procedure that depends on the choice of parameters like the perplexity (\cite{maaten2008visualizing, wattenberg2016how}). 
We therefore interpret these plots only qualitatively and infer that associative domain adaptation learns consistent embeddings for source and target domain that cluster well with observable margins.

\subsubsection{Quantitative evaluation: MMD values}
While t-SNE plots provide qualitative insights into the latent feature representation of a learning algorithm, we want to complement this with a quantitative evaluation and compute the discrepancy in embedding space for target and source domains.
We estimated the MMD with a Gaussian RBF kernel using the TensorFlow implementation provided by the authors of \cite{sutherland2016generative}.

The results are shown in \autoref{tbl:mmd}.
In parentheses we copied the test accuracies on the respective target domains from \autoref{tbl:da}. 

We observe that $\DAmmd$ yields the lowest maximum mean discrepancy, as expected, since this training setup explicitly minimizes this quantity.
At the same time, $\DAassoc$ also reduces this metric in most cases.
Interestingly though, for the setup SVHN $\rightarrow$ MNIST, we actually obtain a particularly high MMD. Nevertheless, the test error of the network trained with $\DAassoc$ is one of the best results.
We ascribe this to the fact that MMD enforces domain invariant feature representations regardless of the source labels, whereas $\Lassoc$ takes into account the labels of associated source samples, resulting in better separation of the clusters and higher similarity within the same class.
Consequently, $\DAassoc$ achieves lower test error on the target domain, which is the actual goal of domain adaptation.

\section{Conclusion} 
We have introduced a novel, intuitive domain adaptation scheme for neural networks termed \emph{associative domain adaptation} that generalizes a recent approach for semi-supervised learning\cite{Haeusser2017} to the domain adaptation setting. 
The key idea is to optimize a joint loss function combining the classification loss on the source domain with an association loss that imposes consistency of source and target embeddings.
The implementation is simple, works with arbitrary architectures in an end-to-end manner and introduces no significant additional computational and structural complexity. 
We have demonstrated the capabilities of associative domain adaptation on various benchmarks and achieved state-of-the-art results for all our experiments. 
Finally, we quantitatively and qualitatively examined how well our approach reduces the discrepancy between network embeddings from the source and target domain. 
We have observed that, compared to explicitly modelling the maximum mean discrepancy as a cost function, the proposed association loss results in embeddings that are more effective for classification in the target domain, the actual goal of domain adaptation.

{\small
\bibliographystyle{ieee.bst}
\bibliography{iccv2017}
}

\clearpage
\onecolumn
   \newpage
   \null
   \vskip .375in
   \begin{center}
      {\Large \bf Supplementary Material for `Associative Domain Adaptation' \par}
      \vspace*{24pt}
      {
      \large
      \lineskip .5em
      \begin{tabular}[t]{c}
          
      \end{tabular}
      \par
      }
      \vskip .5em
      \vspace*{12pt}
   \end{center}

\setcounter{section}{0}
\setcounter{figure}{0}
\setcounter{table}{0}
\setcounter{footnote}{0}

\renewcommand*{\theHsection}{A\thesection}
\renewcommand*{\theHfigure}{A\thefigure}
\renewcommand*{\theHtable}{A\thetable}
\graphicspath{{./supplementary_figures/}}

We provide additional information that is necessary to reproduce our results, as well as plots complementing the evaluation section of the main paper.
To this end, we begin by stating implementation details for our neural network learning algorithm.
Furthermore, we show additional t-SNE embeddings of source target domain for the different domain adaptation tasks analyzed in the paper.

\section{Hyperparameters}
We report the hyperparameters that we used for our experiments for the sake of reproducibility as detailed in \autoref{tbl:params}.

\begin{table*}[b!]
	\begin{center}
		\begin{tabular}{|c||c|c|c|c|}
			\hline
			\multirow{ 2}{*}{Hyperparameter} & \multicolumn{4}{c|}{Domains (source $\rightarrow$ target) } \\
			& MNIST $\rightarrow$ MNIST-M
			& Syn. Digits $\rightarrow$ SVHN 
			& SVHN $\rightarrow$ MNIST
			& Syn. Signs $\rightarrow$ GTSRB\\
			\hline
			\hline
 New width/height                           & 32                          & -                        & 32                        & -                               \\
Source domain batch size                    & 1000                        & 1000                     & 1000                     & 1032                              \\
 Target domain batch size                   & 1000                        & 1000                     & 1000                      & 1032                            \\
 Learning rate decay steps                  & 9000                        & 9000                     & 9000                      & 9000                            \\
 Visit loss weight                          & 0.6                         & 0.2                      & 0.2                       & 0.1                             \\
 Delay (steps) for $\Lassoc$                        & 500                         & 2000                     & 500                       & 0                               \\

			\hline
		\end{tabular}

	\end{center}
			        \caption{Hyperparameters for our domain adaptation experiments.} 
    	\label{tbl:params}

\end{table*}

\section{t-SNE embeddings}
\begin{figure*}[h!]
	\centering
        \begin{subfigure}{0.3\textwidth}
            \includegraphics[width=\textwidth]{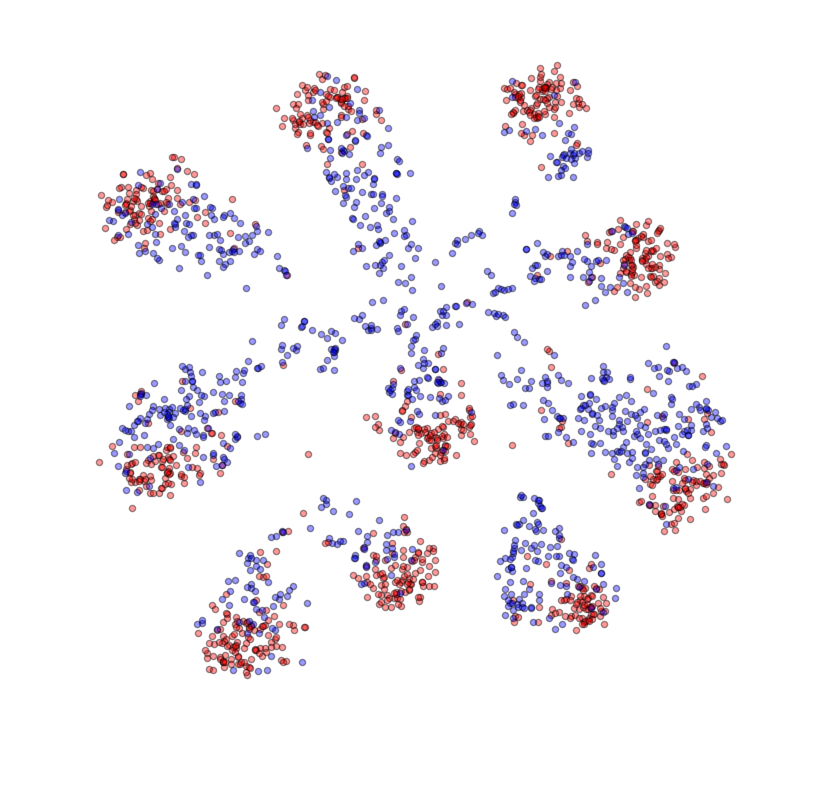}
        \end{subfigure}%
        \begin{subfigure}{0.3\textwidth}
            \includegraphics[width=\textwidth]{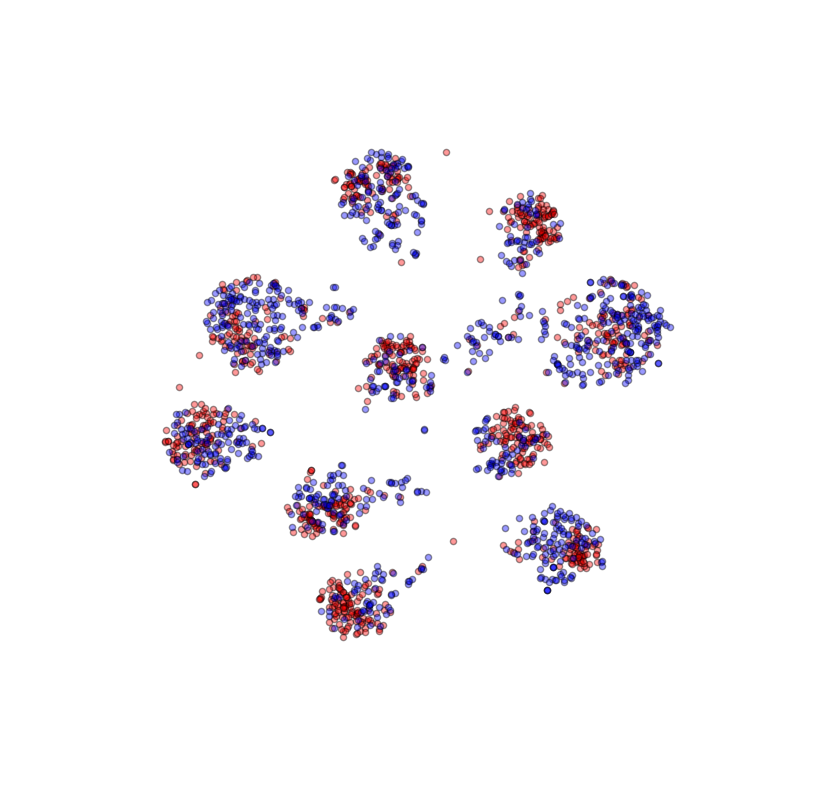}
        \end{subfigure}%
        \begin{subfigure}{0.3\textwidth}
            \includegraphics[width=\textwidth]{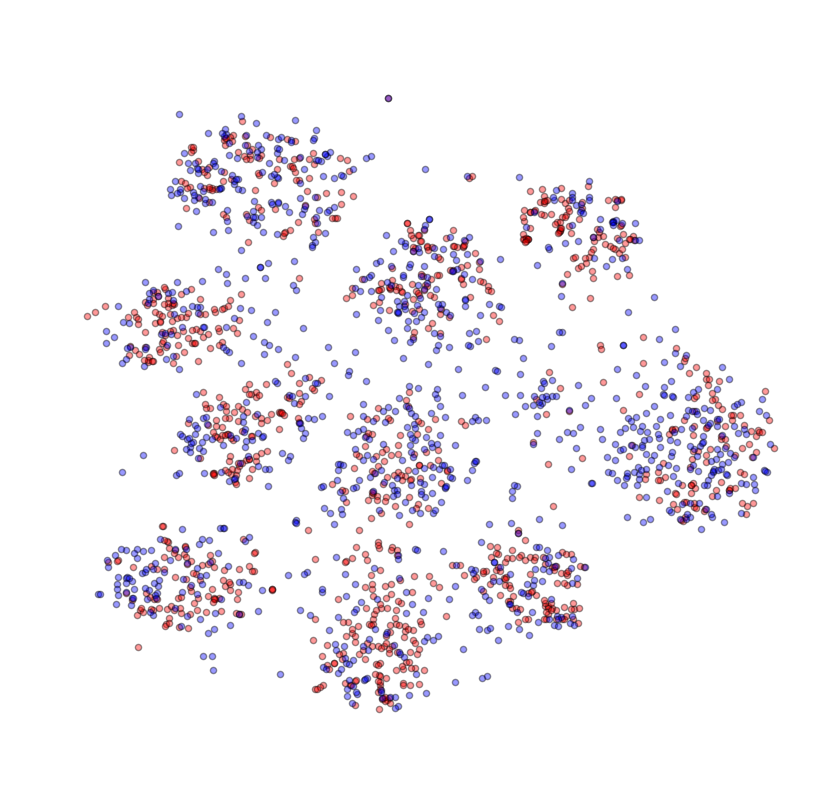}
        \end{subfigure}%
        \\
    	\begin{subfigure}{0.3\textwidth}
            \includegraphics[width=\textwidth]{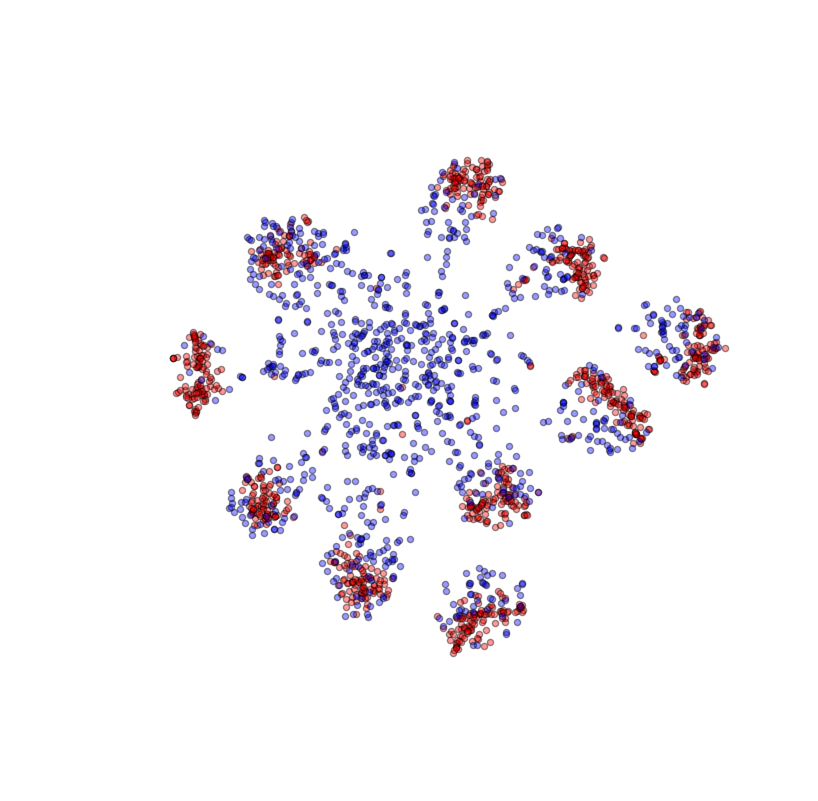}
        \end{subfigure}%
        \begin{subfigure}{0.3\textwidth}
            \includegraphics[width=\textwidth]{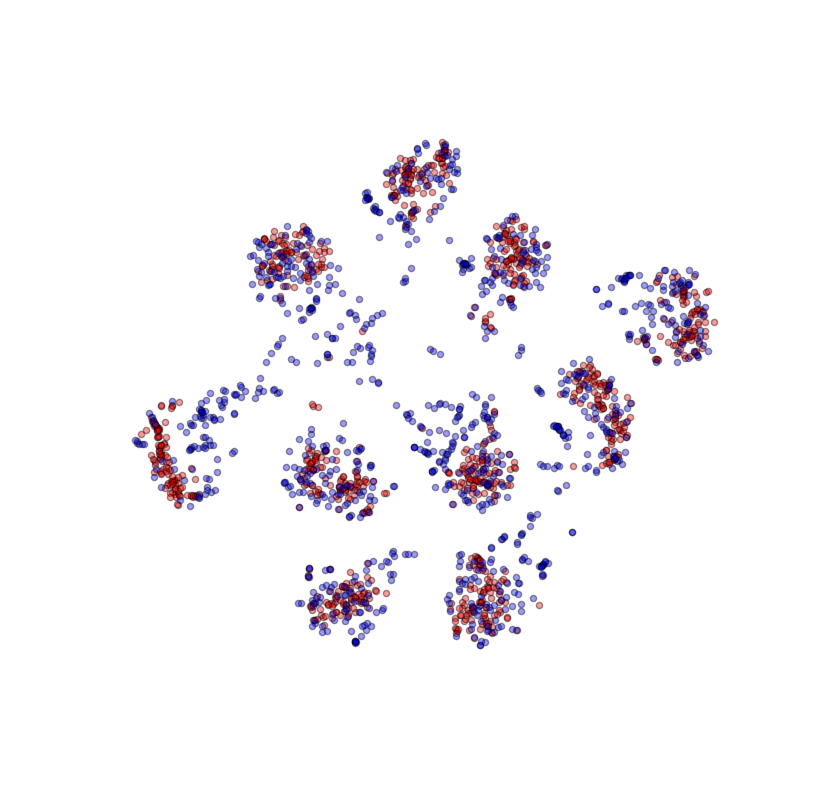}
        \end{subfigure}%
        \begin{subfigure}{0.3\textwidth}
            \includegraphics[width=\textwidth]{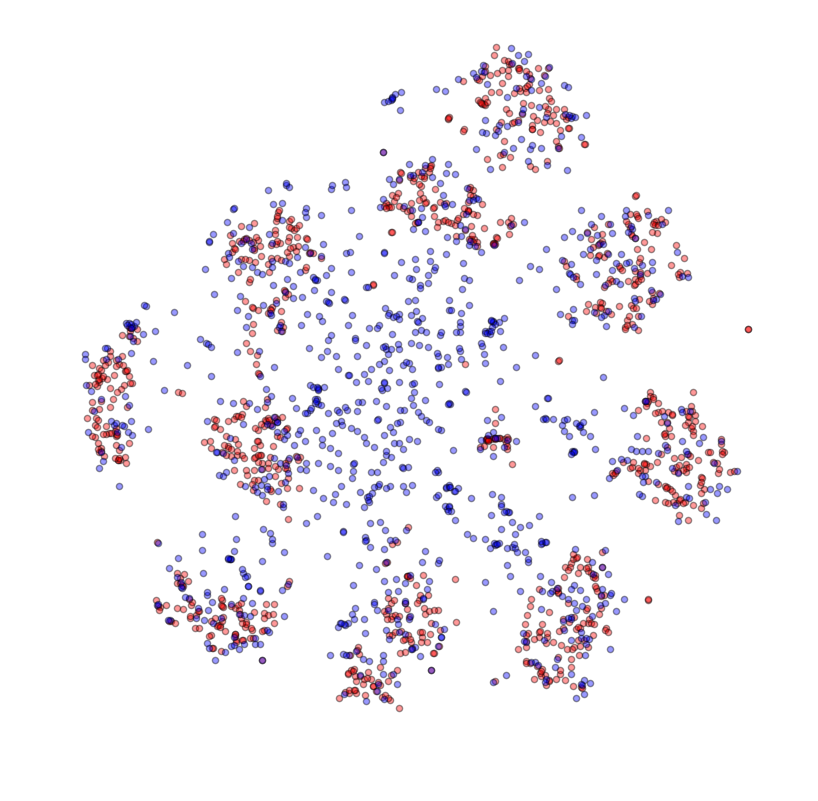}
        \end{subfigure}%
        \\
        \begin{subfigure}{0.3\textwidth}
            \includegraphics[width=\textwidth]{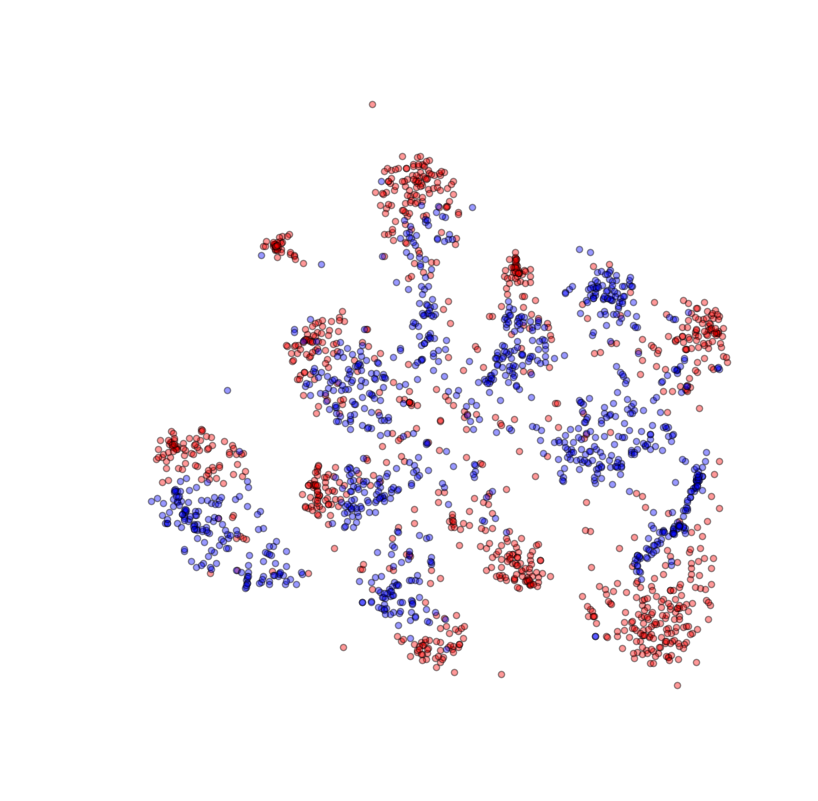}
        \end{subfigure}%
        \begin{subfigure}{0.3\textwidth}
            \includegraphics[width=\textwidth]{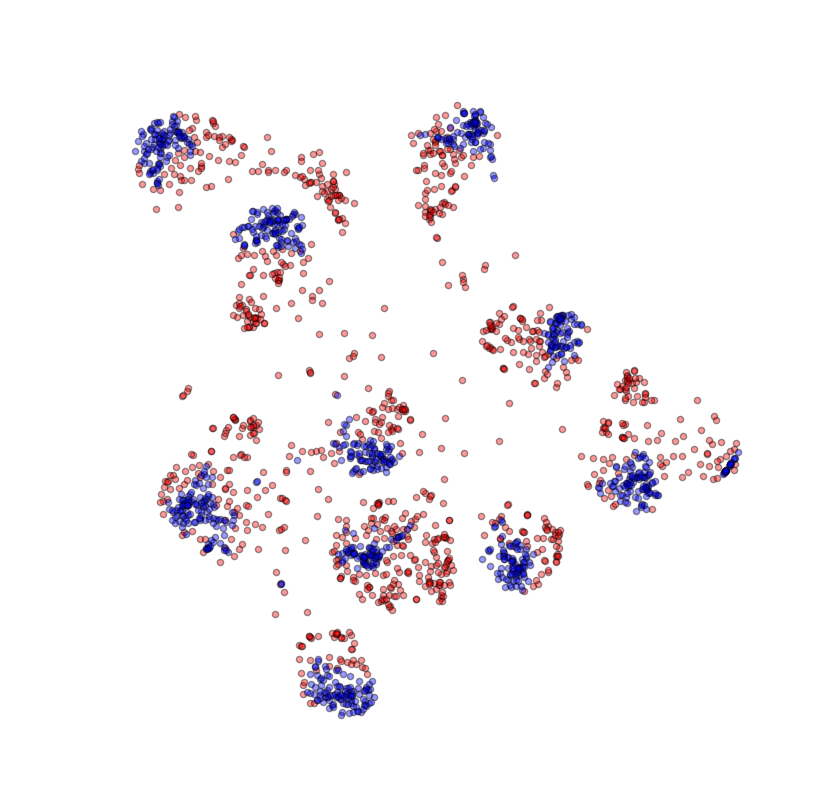}
        \end{subfigure}%
        \begin{subfigure}{0.3\textwidth}
            \includegraphics[width=\textwidth]{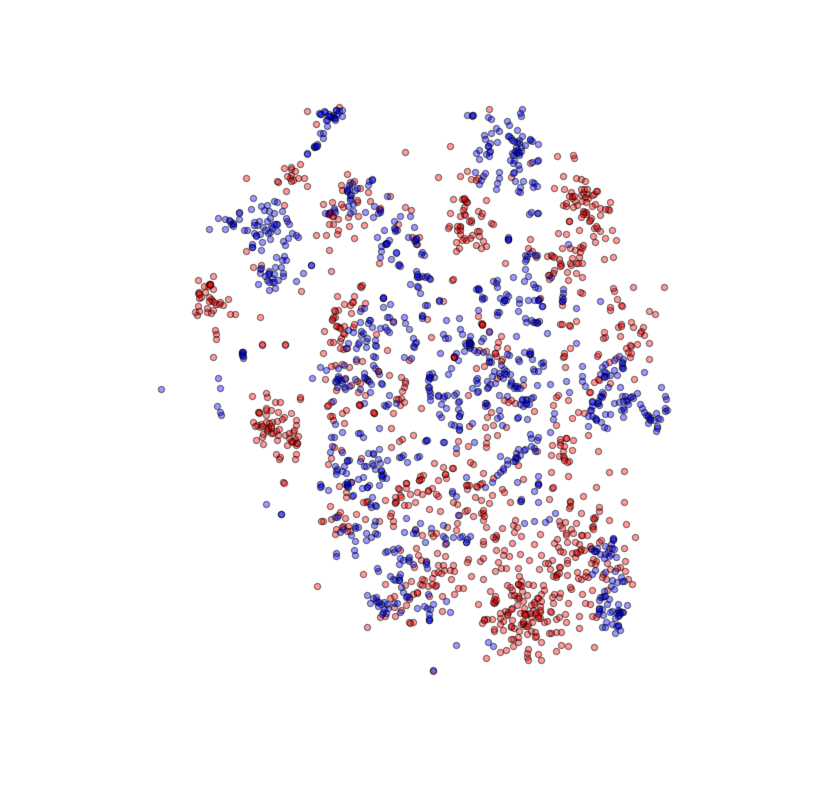}
        \end{subfigure}%
        \\
        \begin{subfigure}{0.3\textwidth}
            \includegraphics[width=\textwidth]{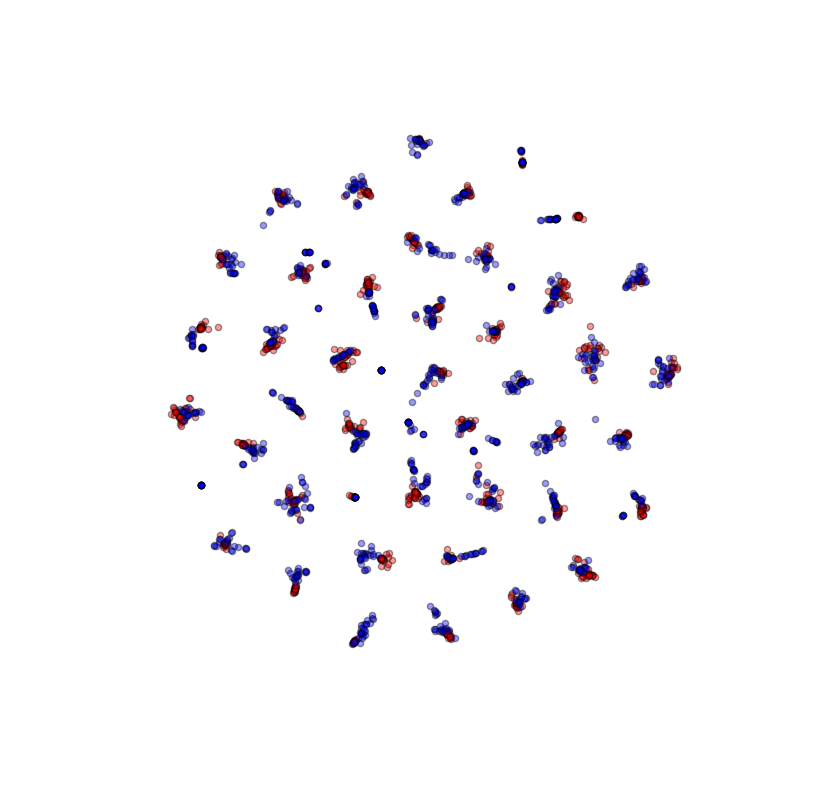}
        \end{subfigure}%
        \begin{subfigure}{0.3\textwidth}
            \includegraphics[width=\textwidth]{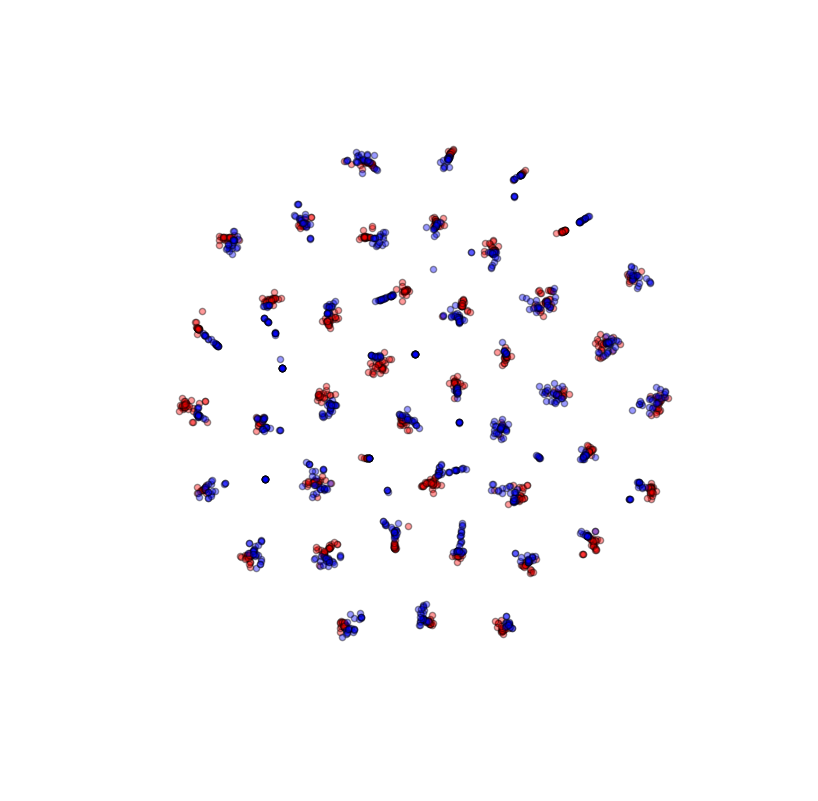}
        \end{subfigure}%
        \begin{subfigure}{0.3\textwidth}
            \includegraphics[width=\textwidth]{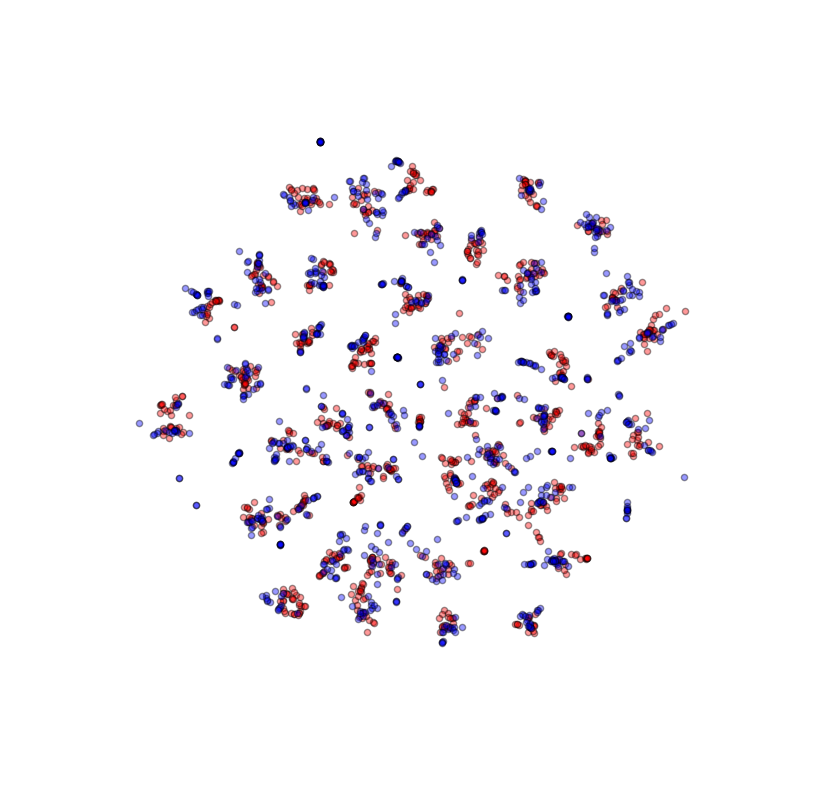}
        \end{subfigure}%
        \caption{t-SNE embeddings of test samples for source (red) and target (blue).
        \textbf{First row:} MNIST to MNIST-M, perplexity $35$.
        \textbf{Second row:} SVHN to MNIST, perplexity $35$.
        \textbf{Third row:} Synthetic Signs to GTSRB, perplexity $25$.
        1,000 samples per domain, except for Synthetic Signs to GTSRB, where we took 60 samples for each of the 43 classes due to class imbalance in GTSRB.
    	\textbf{Left}: After training on \emph{source only}. \textbf{Middle}: after training with \emph{associative domain adaptation} ($\DAassoc$). \textbf{Right}: after training with \emph{MMD loss} ($\DAmmd$).} 
    	\label{fig:supplementary_tsne}
\end{figure*}

We complement our analysis in Section 3.4.1 of the main document, \textit{Qualitative evaluation: t-SNE embeddings}. 
In \autoref{fig:supplementary_tsne} we show the t-SNE embeddings for all domain adaptation tasks that we have analyzed (cf. Table 3 of the main paper).
The qualitative interpretation that we provide for the task Synthetic Digits to SVHN in the main paper is consistent across all tasks: when trained on source only, the target domain distribution is diffuse, the respective target classes can be visibly separated after domain adaptation and the separation is less clear when training with an MMD loss instead of our associative loss.
Note that for the task Synthetic Signs to GTSRB, the target domain test error for the network trained on source only is already rather low. Subsequent domain adaptation improves the numerical result, which is, however, difficult to observe qualitatively due to the relatively small \emph{coverage} compared to the previous settings.

\end{document}